\documentclass[conference]{IEEEtran}
\usepackage{cite}
\usepackage{amsmath,amssymb,amsfonts}
\usepackage{algorithmic}
\usepackage{graphicx}
\usepackage{textcomp}
\usepackage{xcolor}

\usepackage{booktabs}
\usepackage{algorithm}
\usepackage{balance}
\usepackage{epstopdf}

\def\BibTeX{{\rm B\kern-.05em{\sc i\kern-.025em b}\kern-.08em
    T\kern-.1667em\lower.7ex\hbox{E}\kern-.125emX}}
\begin{document}

\title{Robust Position Estimation by Rao-Blackwellized Particle Filter without Integer Ambiguity Resolution in Urban Environments\\
}

\author{\IEEEauthorblockN{1\textsuperscript{st} Daiki Niimi}
\IEEEauthorblockA{\textit{Department of Mechatronics Engineering} \\
\textit{Graduate School of Meijo University}\\
Aichi, Japan \\
243432020@ccmailg.meijo-u.ac.jp}
\and
\IEEEauthorblockN{2\textsuperscript{nd} An Fujino}
\IEEEauthorblockA{\textit{Department of Mechatronics Engineering} \\
\textit{Meijo University}\\
Aichi, Japan \\
210447077@ccmailg.meijo-u.ac.jp}
\and
\IEEEauthorblockN{3\textsuperscript{rd} Taro Suzuki}
\IEEEauthorblockA{\textit{Future Robotics Technology Center} \\
\textit{Chiba Institute of Technology}\\
Chiba, Japan \\
taro@furo.org}
\and
\IEEEauthorblockN{4\textsuperscript{th} Junichi Meguro}
\IEEEauthorblockA{\textit{Department of Mechatronics Engineering} \\
\textit{Meijo University}\\
Aichi, Japan \\
meguro@meijo-u.ac.jp}
}

\maketitle

\begin{abstract}
This study proposes a centimeter-accurate positioning method that utilizes a Rao-Blackwellized particle filter (RBPF) without requiring integer ambiguity resolution in global navigation satellite system (GNSS) carrier phase measurements. The conventional positioning method employing a particle filter (PF) eliminates the necessity for ambiguity resolution by calculating the likelihood from the residuals of the carrier phase based on the particle position. However, this method encounters challenges, particularly in urban environments characterized by non-line-of-sight (NLOS) multipath errors. In such scenarios, PF tracking may fail due to the degradation of velocity estimation accuracy used for state transitions, thereby complicating subsequent position estimation. To address this issue, we apply Rao-Blackwellization to the conventional PF framework, treating position and velocity as distinct states and employing the Kalman filter for velocity estimation. This approach enhances the accuracy of velocity estimation and, consequently, the precision of position estimation. Moreover, the proposed method rejects NLOS multipath signals based on the pseudorange residuals at each particle position during the velocity estimation step. This process not only enhances velocity accuracy, but also preserves particle diversity by allowing particles to transition to unique states with varying velocities. Consequently, particles are more likely to cluster around the true position, thereby enabling more accurate position estimation. Vehicular  experiments in urban environments demonstrated the effectiveness of proposed method in achieving a higher positioning accuracy than conventional PF-based and conventional GNSS positioning methods.
\end{abstract}


\section{Introduction}
The global navigation satellite system (GNSS) is extensively employed as a positioning method for mobile robots and vehicles in outdoor environments. Notably, real-time kinematic (RTK) GNSS, which utilizes double-differenced (DD) carrier phase observations, can achieve centimeter-level position estimation accuracy in real-time by accurately resolving integer ambiguities in the carrier phase observations. Consequently, RTK-GNSS is widely implemented in applications such as mapping and autonomous driving, which require highly accurate position estimation. However, RTK-GNSS faces challenges related to degraded position estimation accuracy when integer ambiguities are incorrectly resolved. Additionally, in urban environments, signal blockages and cycle slips induce discontinuities in the carrier phase, which necessitates the re-resolution of integer ambiguities and complicates the maintenance of continuous, high-precision positioning \cite{gnss_general}.

In contrast, methods that do not require integer ambiguity resolution, yet achieve position estimation accuracy comparable to conventional RTK-GNSS, have been investigated \cite{ar_review}. In a previous study, the authors proposed a method using a particle filter (PF) to determine the probability of particles based on the residuals of the GNSS carrier phase observations \cite{mupf}. This approach facilitated centimeter-level positioning accuracy without the need for integer ambiguity resolution. Nonetheless, the effectiveness of this method is contingent upon the performance of the particle state transition, which ultimately affects its positioning accuracy. If the particles fail to transition to the proximity of the true position during state transitions, the probability cannot be accurately calculated. This method employs the velocity estimated by the least-squares method derived from GNSS Doppler observations for state transitions. However, urban environments introduce significant multipath errors into velocity estimation \cite{vel_multipath}. Notably, non-line-of-sight (NLOS) multipath signals prevalent in urban settings represent a major source of error in GNSS positioning \cite{nlos_general1}. Furthermore, when GNSS signals are nearly entirely obstructed, rendering velocity computation impossible, the performance of the PF significantly degrades.

\begin{figure}[t]
  \centering
  \includegraphics[width=85mm]{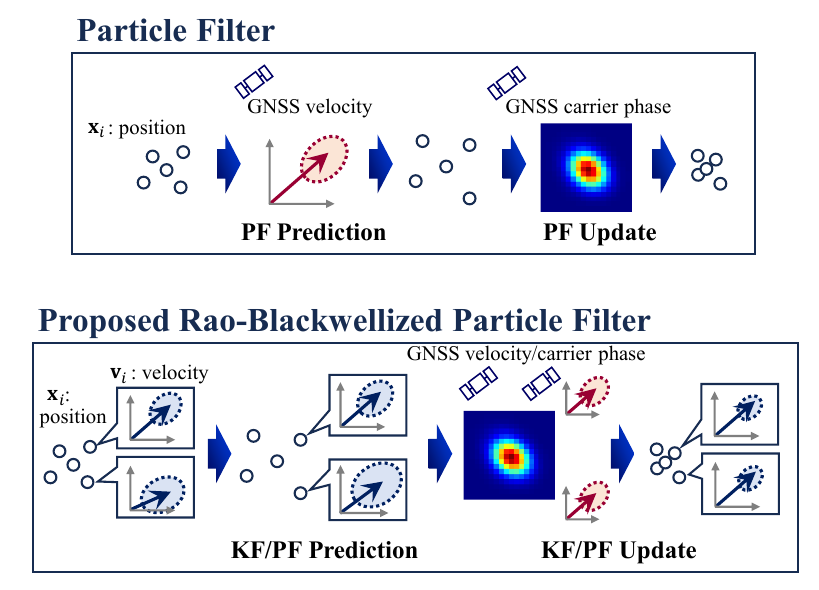}
  \caption{Overview of proposed Rao-Blackwellized PF. Top: Position estimation using conventional PF. The state of PF is the 3D position. Bottom: Proposed Rao-Blackwellized PF. In addition to position, 3D velocity is added to the state, and each particle has its own velocity, which is used for particle prediction.}
  \label{fig1}
\end{figure}

Therefore, this study proposes a centimeter-level accurate positioning method utilizing a Rao-Blackwellized particle filter (RBPF) that eliminates the need for integer ambiguity resolution in GNSS carrier phase observations. Fig. \ref{fig1} presents an overview of the proposed method. The previous PF estimation state encompassed a three-dimensional (3D) position, representing a 3D degree-of-freedom state estimation problem; however, the proposed method integrates a 3D velocity into the estimated state. PF is commonly subject to the curse of dimensionality, rendering the increase of the estimated state dimension impractical. Therefore, we propose extending the conventional PF to an RBPF. By implementing Rao-Blackwellization, the dimensionality of the PF remains unchanged, thus enabling velocity estimation for each particle through a Kalman filter (KF) in conjunction with position estimation via the PF. Furthermore, in the velocity estimation step, the proposed method rejects the NLOS multipath signals based on the pseudorange residuals at each particle position. The accuracy of state transition in the PF is enhanced, facilitating a more effective distribution of particles around the true position. This enhancement is anticipated to improve the accuracy of position estimation in challenging environments such as urban areas.

\section{Related Studies}
\subsection{High-Precision Positioning using GNSS}
RTK-GNSS represents a method that achieves centimeter-level positioning accuracy in real-time by estimating the integer ambiguities contained in DD carrier phase observations \cite{gnss_general}. In standard RTK-GNSS, ambiguities are initially estimated as floating points from DD pseudorange and carrier phase observations utilizing a KF. Subsequently, these ambiguities are refined into integers using an integer least-squares algorithm \cite{ar_review} (e.g., the LAMBDA method \cite{lambda}) to achieve high-precision positioning. However, in urban environments, the existence of NLOS multipath effects can hinder integer ambiguity resolution or lead to incorrect integer ambiguity resolution, resulting in significant degradation of positioning accuracy \cite{rtk_reliability}. Various methods for detecting NLOS multipath have been proposed \cite{nlos_general1,nlos_general2}; however, it remains an open question whether complete rejection of NLOS multipath using GNSS observations alone is feasible. Furthermore, when GNSS signals are obstructed or cycle slips occur, causing discontinuities in the carrier phase, re-resolving the integer ambiguities becomes necessary, which is an additional challenge.

Alternatively, ambiguity function method (AFM) \cite{afm} and its improvement, modified AFM (MAFM) \cite{mafm1,mafm2,mafm3}, which focus solely on the fractional part of the carrier phase, are proposed to estimate the integer ambiguity. AFM is a method that searches for integer ambiguities in a 3D space using only the fractional part (less than one wavelength) of the carrier phase. However, AFM exhibits poor computational efficiency in searching for integer ambiguity within a 3D space, rendering it less practical than integer least-square methods. In our previous study, we devised a method that combines AFM with PF to estimate the likelihood of particles in 3D space, utilizing the ambiguity function value (AFV), which is the objective function in the optimization process for AFM \cite{mupf}. Rather than employing AFV to estimate integer ambiguity, we proposed a method to ascertain particle likelihood through AFV. Nevertheless, as mentioned earlier, this method relies on the particle positions themselves to compute their likelihood. If the particle distribution does not transition in a manner that encompasses the true position, PF is unable to track the position.

\subsection{Rao-Blackwellized Particle Filter}
RBPF, or marginalized particle filter, is a method that disaggregates the state space by factorizing the probability distribution of the state \cite{rbpf1}. Typically, states characterized by nonlinear models are estimated using PF, whereas states modeled linearly are estimated using other filters, such as KF \cite{rbpf2}. RBPF is employed in various fields to address state estimation problems \cite{rbpf_survey}. In the field of robotics, RBPF has been widely used in simultaneous localization and mapping (SLAM) to estimate positions and maps simultaneously \cite{slam_general}. The common PF encounters a problem known as the curse of dimensionality, where representing the probability distribution of a high-dimensional state space with a set of particles results in an exponential increase in the number of particles and, consequently, the computational cost, related to dimensionality \cite{pf_general1,pf_general2}. In contrast, RBPF uses the PF only to estimate states governed by strongly nonlinear models, thereby enabling an increase in the number of estimated state variables without a corresponding increase in the dimensionality of the PF. Using RBPF with SLAM, the robot's position is estimated with a PF, while the map is estimated using a KF or a binary Bayes filter \cite{rbpf_slam1,rbpf_slam2}. This approach facilitates the implementation of SLAM with PF.

Conversely, in GNSS-based positioning, the application of RBPF has been used infrequently. This limited use arises because position estimation using GNSS observations typically does not involve highly nonlinear models, thereby diminishing the necessity for employing a PF. In studies combining GNSS with 3D maps \cite{pf_gnss1,pf_gnss2,pf_gnss3}, PF-based methods have been proposed; however, RBPF has not been utilized. Furthermore, methods that integrate states estimated by GNSS and other sensors \cite{rbpf_gnss1,rbpf_gnss2}, such as inertial navigation systems (INS) or cameras, employing RBPF have been proposed. However, RBPF has not yet been applied to methods relying solely on GNSS.

The contributions of this study are delineated as follows.
\begin{itemize}
   \item An algorithm is proposed that simultaneously estimates position and velocity from GNSS observations using RBPF without estimating integer ambiguities.
   \item The performance of velocity estimation is enhanced by applying RBPF and rejecting NLOS multipath effects based on the residuals of the pseudorange at each particle position.
   \item The proposed method is validated using real data collected in urban environments, demonstrating superior velocity and position estimation accuracy compared to conventional RTK-GNSS and PF.
\end{itemize}

\section{Proposed Method}
In this study, we apply Rao-Blackwellization to a PF framework that estimates position using the likelihood determined by the AFV from the GNSS DD carrier phase observation. By estimating velocity from the GNSS Doppler observation using the KF for each particle and incorporating this estimate into the state transition of each particle, we improve the position estimation accuracy in urban environments.

\subsection{Rao-Blackwellization}
Among the states $\mathbf{x}_t$ to be estimated at time $t$, the nonlinear states are denoted as $\mathbf{x}_t^n$ and the linear states as $\mathbf{x}_t^l$. Given the observation $\mathbf{y}_t$, the factorization of the state probability $p\left(\mathbf{x}t \mid \mathbf{y}{0 \colon t}\right)$ through Rao-Blackwellization is represented as follows.

\begin{equation}
  \begin{aligned}
  p\left(\mathbf{x}_t \mid \mathbf{y}_{0 \colon t}\right) 
  & = p\left(\mathbf{x}_t^l, \mathbf{x}_{0 \colon t}^n \mid \mathbf{y}_{0 \colon t}\right) \\ 
  & = p\left(\mathbf{x}_{0 \colon t}^n \mid \mathbf{y}_{0 \colon t}\right) p\left(\mathbf{x}_t^l \mid \mathbf{x}_{0 \colon t}^n, \mathbf{y}_{0 \colon t}\right)
  \label{eq1}
  \end{aligned}
\end{equation}

\noindent As shown in \eqref{eq1}, Rao-Blackwellization allows the separation of state space. The probability of states with nonlinear models, $p\left(\mathbf{x}_{0 \colon t}^n \mid \mathbf{y}_{0 \colon t}\right)$, can be estimated using a PF. Conversely, the probability of states with linear models, $p\left(\mathbf{x}_t^l \mid \mathbf{x}_{0 \colon t}^n, \mathbf{y}_{0 \colon t}\right)$, can be estimated using filters such as a KF.

In this study, the 3D position is defined as the state $\mathbf{x}_t^n$ with a nonlinear model, while the 3D velocity is defined as the state $\mathbf{x}_t^l$ with a linear model. When the system state is expressed as $\mathbf{x}_t=\begin{bmatrix} \mathbf{x}_t^n & \mathbf{x}_t^l \end{bmatrix}^T$ and the observations (3D position and velocity) are set as $\mathbf{y}_t=\begin{bmatrix} \mathbf{z}_t & \dot{\mathbf{z}}_t \end{bmatrix}^T$, the state equation and the observation equation are formulated as follows.

\begin{equation}
  \mathbf{x}_{t+1} =
  \begin{bmatrix}
    \mathbf{I}_3  &  \Delta t \mathbf{I}_3\\
    \mathbf{O}_3  &  \mathbf{I}_3
  \end{bmatrix}
  \mathbf{x}_t+\mathbf{w}_t
  \label{eq2}
\end{equation}

\begin{equation}
  \mathbf{y}_t = {h}(\mathbf{x}_t)+\mathbf{e}_t
  \label{eq3}
\end{equation}

\noindent $\Delta t$ represents the time step, $\mathbf{w}_t$ and $\mathbf{e}_t$ represent the process and observation noise, respectively, and $h(\mathbf{x}_t)$ denotes the observation model. Separating \eqref{eq2} and \eqref{eq3} into the nonlinear state $\mathbf{x}_t^n$ and the linear state $\mathbf{x}_t^l$, they can be rewritten accordingly.

\begin{equation}
  \mathbf{x}_{t+1}^n = \mathbf{x}_t^n + \Delta t \mathbf{x}_t^l+\mathbf{w}_t^n
  \label{eq4}
\end{equation}

\begin{equation}
  \mathbf{x}_{t+1}^l = \mathbf{x}_t^l+\mathbf{w}_t^l
  \label{eq5}
\end{equation}

\begin{equation}
  \mathbf{y}_t = {h}(\mathbf{x}_t^n)+\mathbf{C}\mathbf{x}_t^l+\mathbf{e}_t
  \label{eq6}
\end{equation}

\noindent Here, $\mathbf{C}$ denotes the linear observation matrix. In the proposed method, the position is updated using \eqref{eq4} with a PF, while the velocity is updated using \eqref{eq5} with a KF. As shown in \eqref{eq6}, the utilization of RBPF allows the position $\mathbf{x}_t^n$ to influence the observation update of the velocity $\mathbf{x}_t^l$. This arrangement facilitates continuous velocity estimation even in the absence of velocity observations. The algorithm \ref{alg1} presents the RBPF algorithm proposed in this study. The key components of the proposed algorithm are the likelihood determination method for the PF and the velocity estimation method using the KF.

\begin{algorithm}[t]
    \caption{\\ Proposed Rao-Blackwellized Particle Filter}
    \label{alg1}
    \begin{algorithmic}
    \WHILE{$epoch < end\_epoch$}
    \IF{$epoch == 1$}
    \STATE 1) Initialization:For each particle
    \\ \hspace{0.3cm} PF Initialization:
    \\ \hspace{0.7cm} Scatter particles based on any distribution.
    \\ \hspace{0.3cm} KF Initialization:
    \\ \hspace{0.7cm} Set initial velocity and covariance matrix 
    \\ \hspace{0.7cm} for each particle.
    \ELSIF{$epoch > 1$}
    \STATE 2) PF prediction
    \\ \hspace{0.3cm} State transition using KF estimated velocity.
    \STATE 3) KF time update \hspace{0.1cm} Eq.(16) $\sim$ (19)
    \\ \hspace{0.3cm} Update velocity using each particle position.
    \ENDIF  
    \STATE 4) PF correction \hspace{0.1cm} Eq.(7) $\sim$ (10)
    \\ \hspace{0.3cm}Calculate likelihood (AFV) / Resampling particles
    \STATE 5) Calculate NLOS-rejected Doppler velocity
    \STATE 6) KF measurement update \hspace{0.1cm} Eq.(11) $\sim$ (14)
    \\ \hspace{0.3cm} Update velocity using AFV and Doppler velocity.
    \ENDWHILE
    \end{algorithmic}
\end{algorithm}

\subsection{Likelihood Determination for PF}
Notably, the likelihood determination approach for the PF incorporates the method utilized in our previous research \cite{mupf}. Specifically, the likelihood of the particles is determined from the DD residuals of the GNSS observations (pseudorange and carrier phase) without resolving integer ambiguities.

As part of the likelihood determination process, the DD GNSS observations are initially calculated using GNSS measurements from the rover and base stations. The GNSS pseudorange provides information free from ambiguity; therefore, the DD pseudorange consists solely of the DD geometric distance between the satellite and the receiver. In this context, the DD pseudorange for the $k$-th satellite is denoted as $\rho^k$ and the DD geometric distance between the satellite and the $i$-th particle is represented as $r^k(\mathbf{x}_i^n)$. The DD pseudorange residual $d(\rho^k, \mathbf{x}_i^n)$ at the 3D position $\mathbf{x}_i^n$ of the $i$-th particle can be calculated as expressed in \eqref{eq7}.

\begin{equation}
  d(\rho^k, \mathbf{x}_i^n) = \rho^k - r^k(\mathbf{x}_i^n)
  \label{eq7}
\end{equation}

\noindent This DD pseudorange residual encompasses multipath error and pseudorange observation noise.

In contrast, the GNSS carrier phase includes integer ambiguities; therefore, the DD carrier phase incorporates not only the term of the DD geometric distance between the satellite and the receiver but also the term of the DD integer ambiguities. Consequently, resolving the integer ambiguities is essential. However, integer ambiguity resolution is associated with problems such as the cycle slip and re-resolving integer ambiguities. Therefore, in the proposed method, the AFV, which is utilized as the objective function of the AFM \cite{afm}, is employed to determine the likelihood in the PF. The AFV allows the likelihood of particles to be estimated without integer ambiguity resolution.

When the DD carrier phase for the $k$-th satellite is represented as $\Phi^k$, the value of the AFV at the 3D position $\mathbf{x}_i^n$ of the $i$-th particle, denoted as $\psi\left(\Phi^k, \mathbf{x}_i^n\right)$ is expressed by the following equation \cite{afm}.

\begin{equation}
  \psi\left(\Phi^k, \mathbf{x}_i^n \right)=\operatorname{round}\left(\Phi^k-\frac{1}{\lambda} r^k\left(\mathbf{x}_i^n \right)\right)-\left(\Phi^k-\frac{1}{\lambda} r^k\left(\mathbf{x}_i^n \right)\right)
  \label{eq8}
\end{equation}

\noindent Here, $\operatorname{round}$ refers to the rounding function, and $\lambda$ represents the wavelength of the carrier phase. In \eqref{eq8}, the influence of integer ambiguities is eliminated, thus focusing exclusively on the fractional part of the carrier phase, which remains less than one wavelength. The AFV value becomes zero when the particle $\mathbf{x}_i^n$ corresponds to the true position, allowing the likelihood of the $i$-th particle for the $k$-th satellite to be calculated using the following equation:

\begin{equation}
  p\left(\Phi^k \mid \mathbf{x}_i^n \right)=\frac{1}{\sqrt{2 \pi} \sigma_\Phi} \exp \left(-\frac{\psi\left(\Phi^k,\mathbf{x}_i^n \right)^2}{2 \sigma_\Phi^2}\right)
  \label{eq9}
\end{equation}

\noindent where $\sigma_\Phi$ indicates the standard deviation of the observation accuracy for the DD carrier phase. Notably, the AFV value becomes zero at intervals of the carrier wavelength $\lambda$ along the LOS direction of the satellites, resulting in the emergence of multiple likelihood peaks in the 3D space. Consequently, by combining the AFV values obtained from the carrier phase observations of multiple satellites, local peaks can be suppressed. If the total number of satellites with carrier phase observations is $K$, the final likelihood of the particle can be expressed by the following equation.

\begin{equation}
  p\left(\boldsymbol{\Phi} \mid \mathbf{x}_i^n \right)=\prod_{k=1}^K p\left(\Phi^k \mid \mathbf{x}_i^n \right)
  \label{eq10}
\end{equation}

\begin{figure}[t]
  \centering
  \includegraphics[width=88mm]{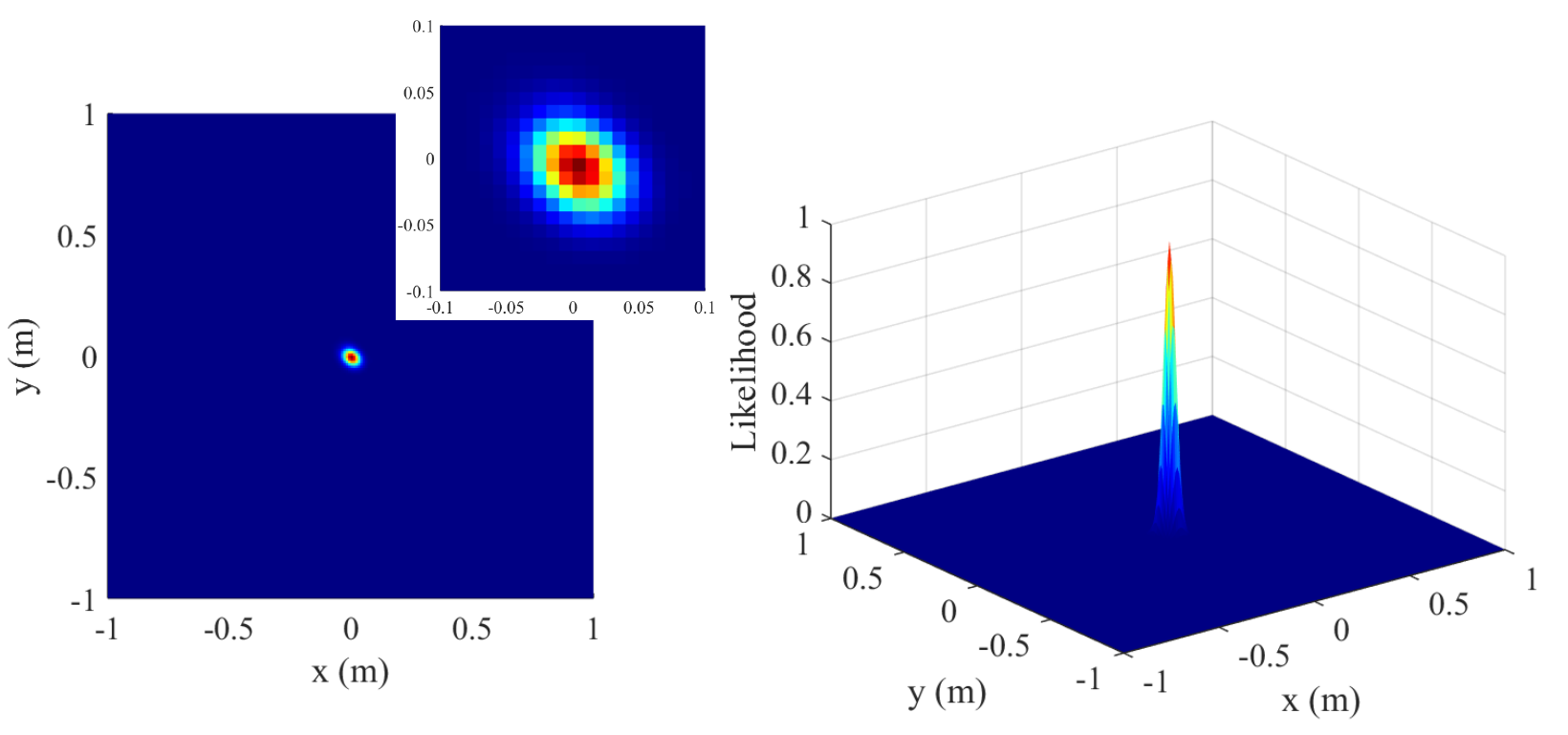}
  \caption{Example of likelihood computed from AFV. The center of the figure is the true position. The likelihood based on AFV has a sharp peak near the true position.}
  \label{fig2}
\end{figure}

Fig.~\ref{fig2} illustrates an example of the likelihood distribution calculated using \eqref{eq10} from GNSS data acquired in an open-sky environment. In this figure, the likelihood was computed from the AFV values at grid points spaced 1 cm apart, with the center representing the true position. As illustrated in Fig.~\ref{fig2}, the likelihood distribution determined from the AFV reveals a single sharp peak at the true position in an environment characterized by minimal multipath effects. This observation demonstrates the potential to estimate the position to the centimeter level without estimating the ambiguity estimation using AFV.

From \eqref{eq9}, it is evident that determining the likelihood of a particle using the AFV requires the position of the particles themselves. This requirement indicates that such an approach is not feasible with a KF but can be realized using a PF, which approximates the state (position) as a set of particles. Furthermore, as shown in Fig.~\ref{fig2}, the likelihood based on AFV exhibits sharp peaks. If particles do not transition to near the true position during the state transition in the PF, the PF can collapse. Consequently, enhancing the accuracy of state transitions (velocity estimation performance) becomes crucial.

\subsection{Velocity Estimation using KF}
Velocity estimation using KF in RBPF comprises two steps: measurement update and time update.

\subsubsection{Measurement Update Step}
The measurement update of the state $\mathbf{x}_t^l$ and its covariance $\mathbf{P}_t$ are represented by the following equations.

\begin{equation}
  \mathbf{\hat{x}}_{t \mid t}^l = \mathbf{\hat{x}}_{t \mid t-1}^l+\mathbf{K}_t\left(\mathbf{y}_t-{h}(\mathbf{x}_t^n)-\mathbf{C} \mathbf{\hat{x}}_{t \mid t-1}^l\right)
  \label{eq11}
\end{equation}

\begin{equation}
  \mathbf{P}_{t \mid t} = \mathbf{P}_{t \mid t-1}-\mathbf{K}_t \mathbf{M}_t \mathbf{K}_t^T 
  \label{eq12}
\end{equation}

\noindent $\mathbf{K}_t$ represents the Kalman gain, while $\mathbf{M}_t$ denotes the covariance matrix of the updated state following the measurement update. These parameters are defined by their corresponding equations.

\begin{equation}
  \mathbf{K}_t=\mathbf{P}_{t \mid t-1} \mathbf{C}^T \mathbf{M}_t^{-1}
  \label{eq13}
\end{equation}

\begin{equation}
  \mathbf{M}_t = \mathbf{C}_t \mathbf{P}_{t \mid t-1} \mathbf{C}^T+\mathbf{R}_t
  \label{eq14}
\end{equation}

\noindent $\mathbf{R}_t$ signifies the covariance matrix of the observation noise. In \eqref{eq11}, the term $\left(\mathbf{y}_t-{h}(\mathbf{x}_t^n)-\mathbf{C} \mathbf{\hat{x}}_{t \mid t-1}^l\right)$ indicates innovation in a standard KF. However, the proposed method calculates this using both linear and nonlinear states, specifically velocity and position. 

For velocity observations, our previous methods based on the PF utilized the 3D velocity computed from GNSS Doppler measurements \cite{mupf}. Nevertheless, applying a single 3D velocity observation to multiple particles results in all particles transitioning to the same velocity. This situation can induce a biased distribution of particles, potentially undermining the fundamental advantage of the PF, which is its capacity to estimate states based on various distinct hypotheses. Consequently, we identify and exclude Doppler measurements from NLOS satellites influenced by multipath errors, relying on the position of the particle $\mathbf{x}_t^n$ and generate a subset of LOS observations for each particle. Thereafter, the 3D velocity is calculated individually for each particle using Doppler observations, and the measurement update of KF is performed for each particle at a different velocity. The DD pseudorange residual at each particle position is utilized to identify NLOS. By applying a threshold $\eta$ to the absolute value of the DD pseudorange residuals calculated using \eqref{eq7}, Doppler measurements from NLOS satellites are excluded.

\begin{equation}
  \operatorname{abs}\left( d(\rho^k, \mathbf{x}_i^n) \right) > \eta
  \label{eq15}
\end{equation}

\noindent The threshold value $\eta$ was determined experimentally. This approach enhances velocity estimation accuracy through NLOS rejection while preserving particle diversity by allowing each particle to transition with a distinct velocity. Consequently, more accurate position estimation is expected.

For position observation, the AFV is employed to evaluate the likelihood based on the particle positions, rendering direct acquisition of position observations unfeasible. Therefore, the AFV value is applied directly to the innovation $\mathbf{y}_t^n-h(\mathbf{x}_t^n)$ of the nonlinear state. The AFV approaches zero when the particle position is close to the true position, resulting in smaller innovations for particles near the true position, while those farther away exhibit larger innovations. Therefore, during the process involving the nonlinear state variables in the KF, particles close to the true position are evaluated appropriately, whereas significant corrections are applied to particles distant from the true position. Overall, this mechanism is expected to lead to more accurate velocity estimation.

\subsubsection{Time Update Step}
The time update of the KF within the framework of the RBPF is represented by the following equation:

\begin{equation}
  \mathbf{x}_{t+1 \mid t}^l = \mathbf{\bar{A}}_t^l \mathbf{\hat{x}}_{t \mid t}^l+\mathbf{L}_t\left(\left(\mathbf{x}_{t+1}^n-\mathbf{x}_{t}^n\right)-\mathbf{A}_t^n \mathbf{\hat{x}}_{t \mid t}^l\right)
  \label{eq16}
\end{equation}

\begin{equation}
  \mathbf{P}_{t+1 \mid t}= \mathbf{\bar{A}}_t^l \mathbf{P}_{t \mid t}\left(\mathbf{\bar{A}}_t^l\right)^T-\mathbf{L}_t \mathbf{N}_t \mathbf{L}_t^T
  \label{eq17}
\end{equation}

\noindent where $\mathbf{A}_t^n$ and $\mathbf{A}_t^l$ denote the state transition matrices for the nonlinear and linear states, respectively. The matrix $\mathbf{L}_t$ reflects the gain associated with changes in position, while $\mathbf{N}_t$ represents the covariance matrix of the time-updated state. These elements are defined by the following equation.

\begin{equation}
  \mathbf{L}_t= \mathbf{\bar{A}}_t^l \mathbf{P}_{t \mid t}\left(\mathbf{A}_t^n\right)^T \mathbf{N}_t^{-1}
  \label{eq18}
\end{equation}

\begin{equation}
  \mathbf{N}_t= \mathbf{\bar{A}}_t^n \mathbf{P}_{t \mid t}\left(\mathbf{A}_t^n\right)^T+ \mathbf{Q}_t^n
  \label{eq19}
\end{equation}

\noindent $\mathbf{Q}_t^n$ signifies the covariance matrix of the state noise. In \eqref{eq16}, the term $\left(\left(\mathbf{x}_{t+1}^n-\mathbf{x}_{t}^n\right)-\mathbf{A}_t^n \mathbf{\hat{x}}_{t \mid t}^l\right)$ computes the innovation based on the positional changes of the particles before and after the state transition, as well as the predicted velocity. This means that the estimated velocity during the KF measurement update step, $\hat{\mathbf{x}}_{t \mid t}^l$, integrates with the velocity derived from the displacement of particles resulting from the state transition in the PF, encapsulating the positional changes in the velocity update. This allows for velocity estimation to be maintained even when velocity observations are not available. Consequently, PF collapse can be prevented, especially in urban environments where GNSS signals are interrupted.

\begin{table}[t]
  \centering
  \caption{Equipment used for evaluation}
  \label{tab1}
  \begin{tabular}{@{}ccccc@{}}
    \toprule
    \textbf{Equipment}    & \textbf{Manufacturer} & \textbf{Model} \\ \hline\hline
    GNSS antenna          & Trimble               & Zephyr model 3 \\
    GNSS receiver         & Septentrio            & mosaic-X5      \\
    Ground truth          & Applanix              & POS LV 220     \\ \bottomrule
  \end{tabular}%
\end{table}

\begin{figure}[t]
  \centering
  \includegraphics[width=80mm]{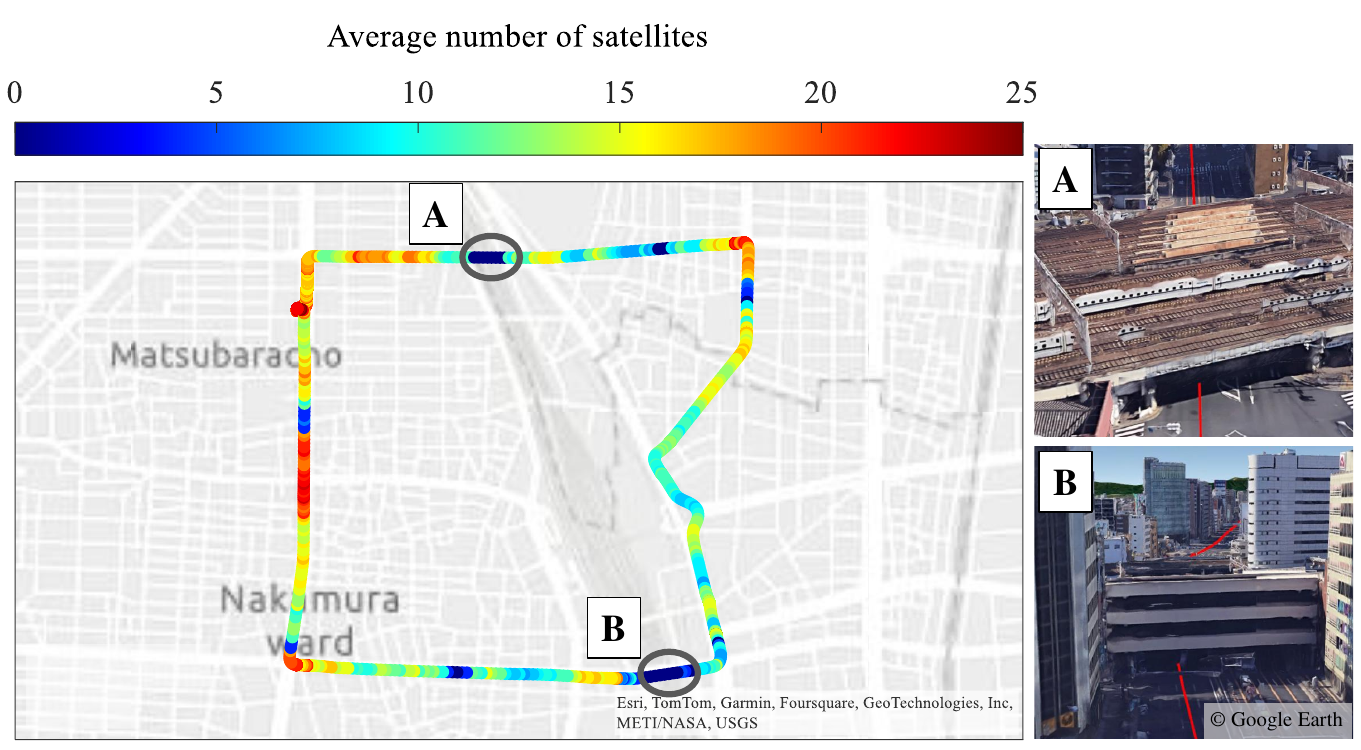}
  \caption{Route and environment of the vehicle for the evaluation test. Colors in the figure indicate the number of receiving satellites and environments where satellites are blocked by overhead structures.}
  \label{fig3}
\end{figure}

\section{Evaluation Test in Urban Environment}
To evaluate the proposed method, we compared the accuracy of velocity and position estimation with a previous PF-based method \cite{mupf} and conventional RTK-GNSS, utilizing GNSS data collected from a vehicle in a real urban environment. The evaluation uses GPS, Galileo, BeiDou and QZSS satellites with a data acquisition interval of 1 Hz. In both the proposed method and the conventional PF, the number of particles was set to 2000. Additionally, for the conventional RTK-GNSS method, we employed RTKLIB \cite{rtklib}, an open source GNSS positioning software package. RTKLIB was configured in kinematic mode, with a 15 $^\circ$ elevation mask and a 35 dB-Hz signal strength mask, while all other parameters were set to their default values.

The evaluation utilized data collected around Nagoya Station in Aichi Prefecture, Japan, using a vehicle equipped with a GNSS antenna. Table~\ref{tab1} presents the equipment used for the evaluation and Fig.~\ref{fig3} illustrates the driving course and environment employed for the evaluation. As ground truth, we used Applanix POS LV 220, a high-grade GNSS, INS and wheel speed sensor integration system. The color of the trajectory in the figure indicates the number of satellites received. In addition, points A and B in the figure indicate underpasses where GNSS signals are completely blocked. 

\begin{figure}[t]
  \centering
  \includegraphics[width=80mm]{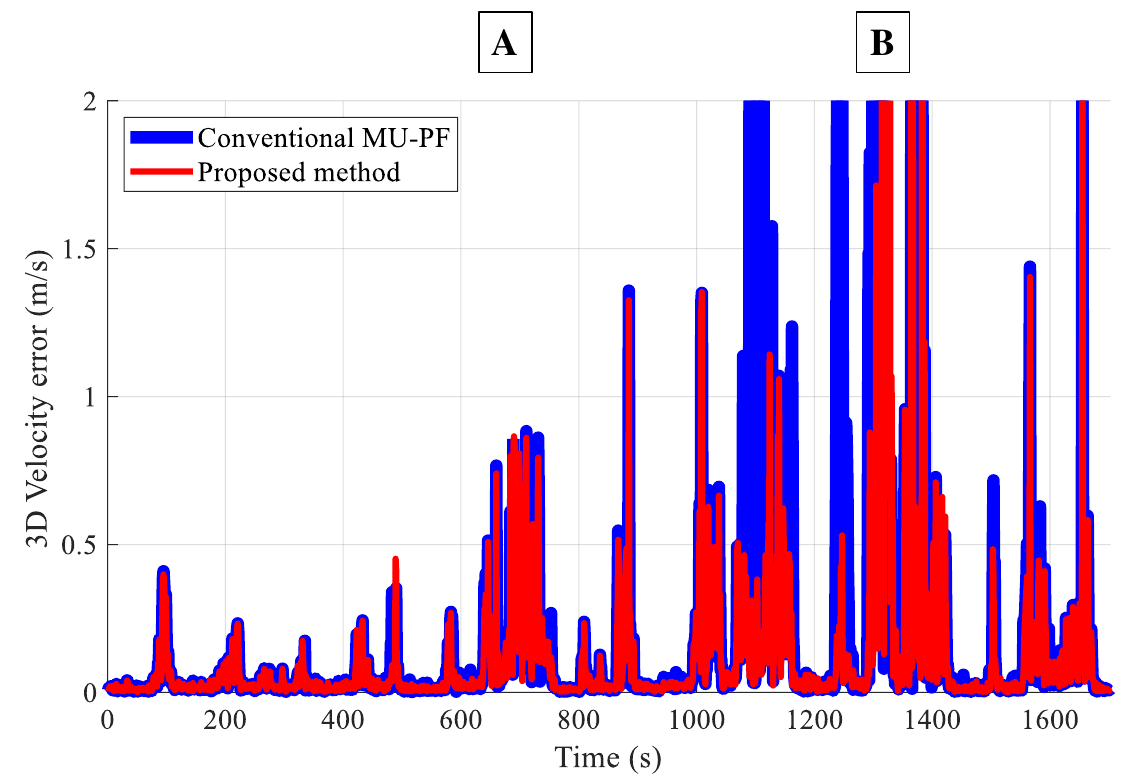}
  \caption{Comparison of the 3D velocity error between the proposed method (red) and conventional PF (blue). The proposed method can calculate velocity even at locations where conventional PF cannot calculate velocity.}
  \label{fig4}
\end{figure}

\begin{figure}[t]
  \centering
  \includegraphics[width=80mm]{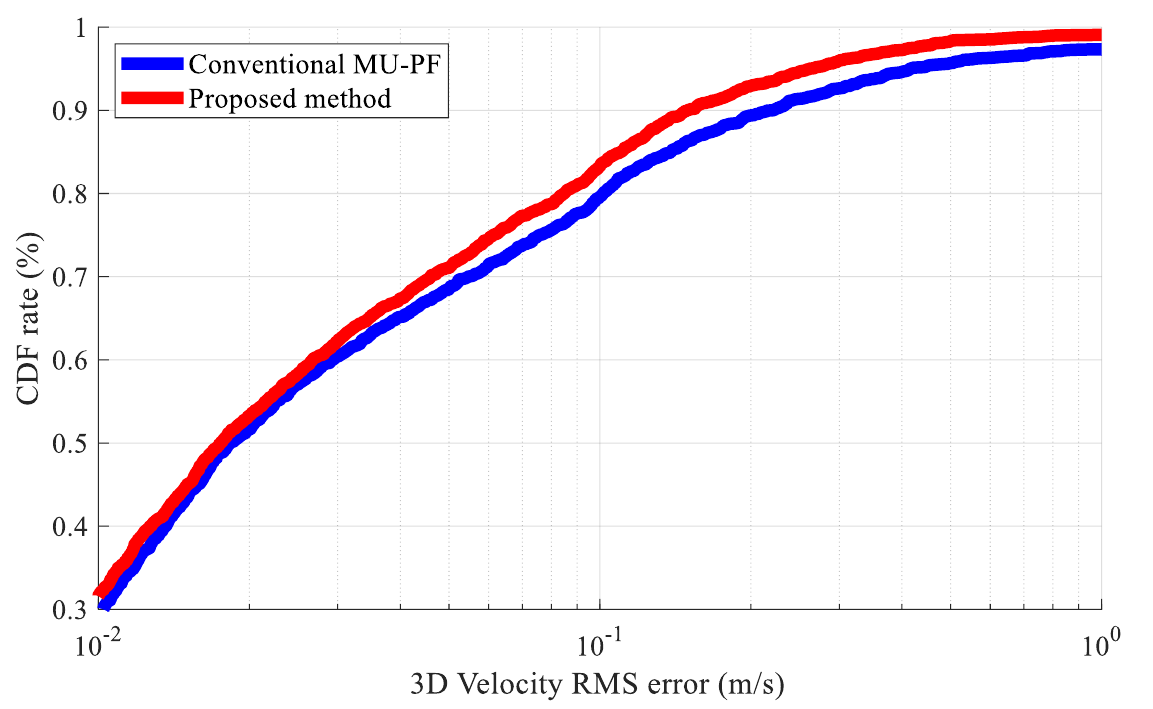}
  \caption{Comparison of the cumulative distribution function of the 3D velocity error. The proposed method (red) has better velocity estimation performance than the conventional PF (blue).}
  \label{fig5}
\end{figure}

\subsection{Evaluation of Velocity Estimation Accuracy}
First, we compare the estimation accuracy of the 3D velocity utilized for state transition between the proposed method and the conventional PF. Since the proposed method estimates different velocity for each particle; thus, the average velocity of all particles is employed for evaluation. Conversely, the velocity in the conventional PF is derived from GNSS Doppler measurements via the least-square method. Fig.~\ref{fig4} shows the 3D velocity errors for both methods. Furthermore, Fig.~\ref{fig5} illustrates the cumulative distribution function (CDF) of the 3D velocity errors.

\begin{figure}[t]
  \centering
  \includegraphics[width=80mm]{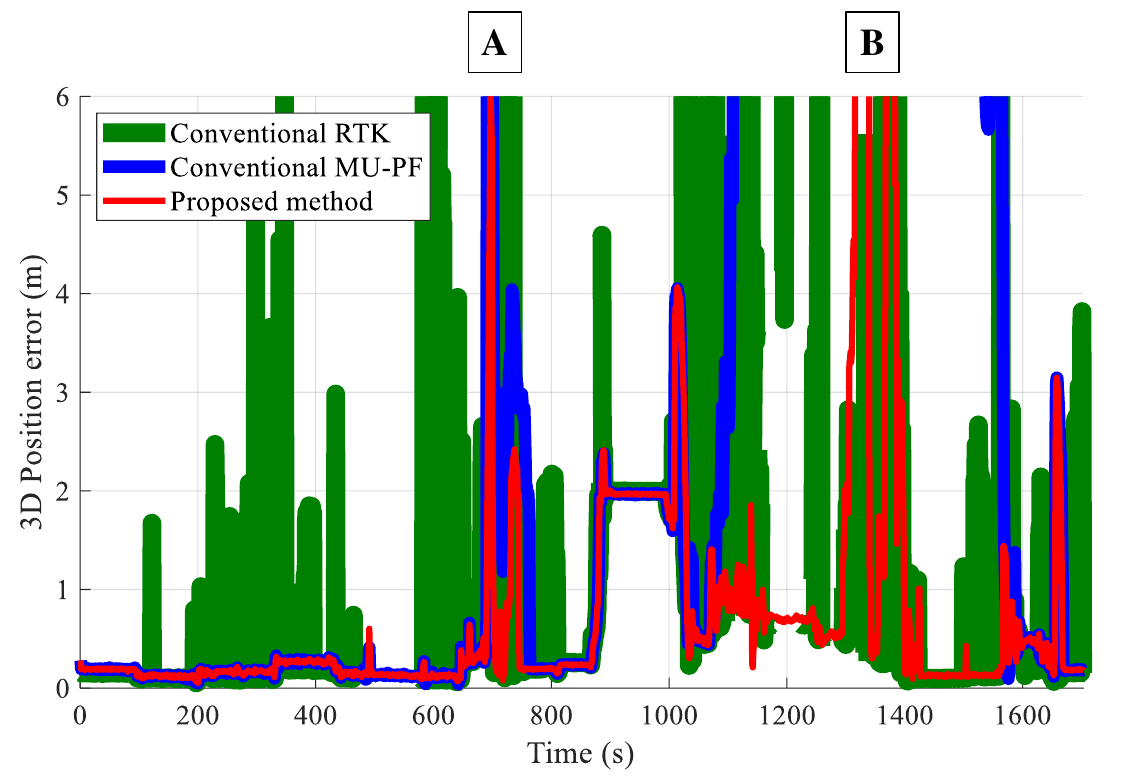}
  \caption{Comparison of the 3D position estimation error of the proposed method (red), the conventional PF (blue), and the conventional RTK-GNSS (green).}
  \label{fig6}
\end{figure}

\begin{figure}[t]
  \centering
  \includegraphics[width=80mm]{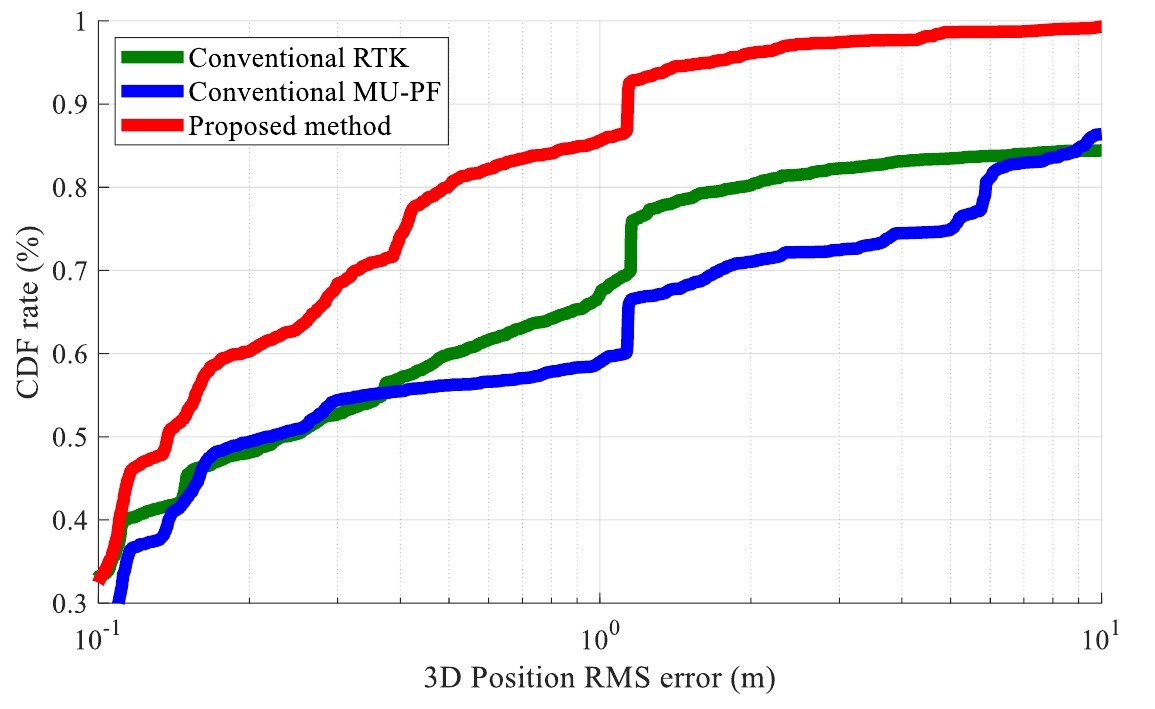}
  \caption{Comparison of the CDF of the position estimation error. The proposed method (red) has better accuracy than the conventional PF (blue) and conventional RTK-GNSS (green).}
  \label{fig7}
\end{figure}

In Fig.~\ref{fig4}, at points A and B, where GNSS signals are entirely obstructed and GNSS observations are unavailable, the conventional PF fails to calculate velocity, resulting in missing values. In contrast, the proposed method updates the velocity using position information during the time update step of the KF. This capability enables the proposed method to sustain velocity estimation even in environments where GNSS observations are rarely accessible. Furthermore, as presented in Fig.~\ref{fig5}, the proportion of velocity estimation accuracy within 0.1 m/s is 79.6\% for the conventional PF, while it improves to 83.3\% with the proposed method. These results demonstrate that the proposed method can estimate velocity stably even in environments where GNSS signals are almost blocked, thus achieving more reliable velocity estimation than the conventional PF.

\subsection{Evaluation of Position Estimation Accuracy}
The accuracy of 3D position estimation is compared among the proposed method, the conventional PF, and the conventional RTK-GNSS \cite{rtklib}. Fig.~\ref{fig6} shows the 3D position errors for the proposed method, conventional PF, and conventional RTK-GNSS. Furthermore, Fig.~\ref{fig7} illustrates the CDF of the 3D position errors. 

In Fig.~\ref{fig6}, at points A and B, where GNSS signals are obstructed, all three methods exhibit an increase in position error or result in missing values. However, the proposed method demonstrates a smaller overall increase in position error than other methods. Additionally, in the conventional PF, the particles cannot be located near the true position, resulting in significant degradation of position estimation. In contrast, the proposed method benefits from enhanced velocity estimation accuracy, which allows for a closer distribution of particles to the true position, thereby reducing instances of severe position estimation degradation. Moreover, as illustrated in Fig.~\ref{fig7}, the proportion of position estimation accuracy within 0.3 m is 54.4\% for the conventional PF and 52.7\% for the conventional RTK-GNSS, while the proposed method improves to 68.5\%.

\subsection{Evaluation of Number of Particles}
We compare the position estimation performance of the proposed method and the conventional PF by varying the number of particles as 1000, 1500, 2000, 2500, and 3000. In this evaluation, a higher number of particles correlates with an increased probability of the particles encompassing the true position. Consequently, a greater particle count is anticipated to enhance position estimation accuracy.

\begin{figure}[t]
  \centering
  \includegraphics[width=80mm]{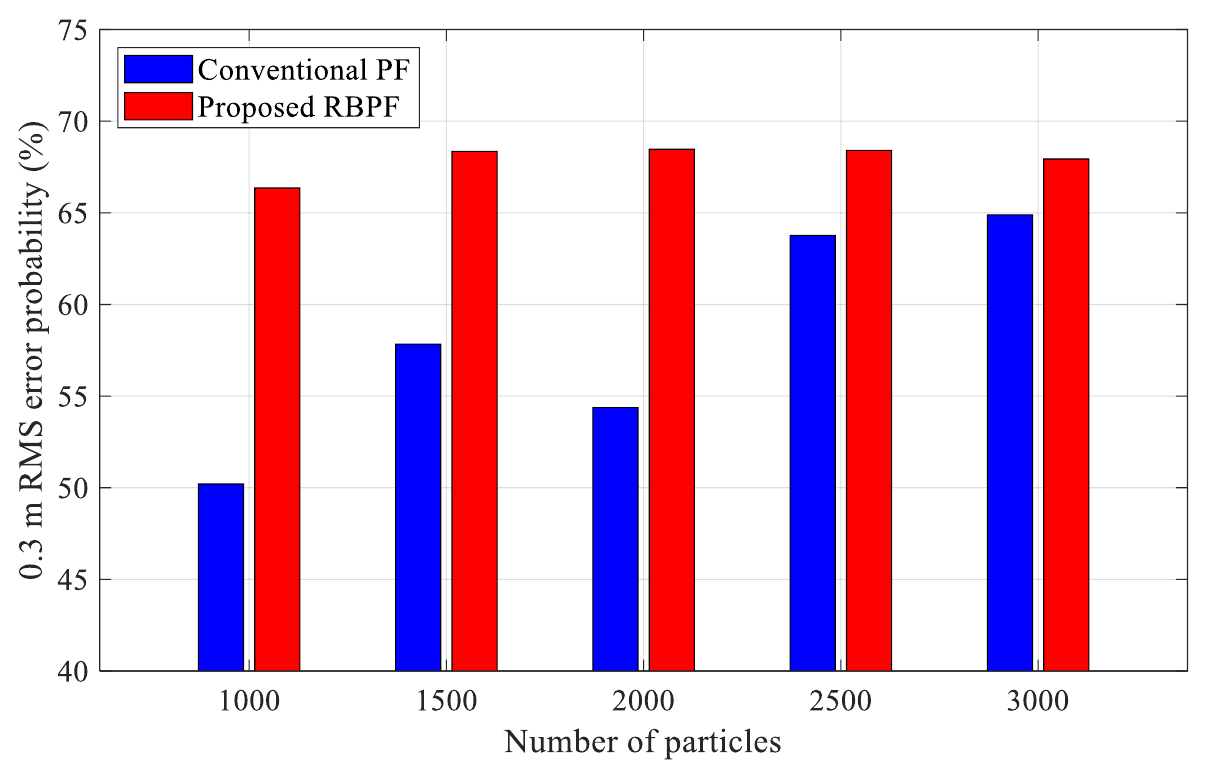}
  \caption{Comparison of the probability of 3D position error as a function of the number of particles between the proposed method (red) and the conventional PF (blue). The proposed method maintains position estimation accuracy even as the number of particles decreases.}
  \label{fig8}
\end{figure}

Fig.~\ref{fig8} illustrates the variation in the probability of position estimation accuracy within 0.3 m for both the proposed method and the conventional PF as the number of particles. From Fig.~\ref{fig8}, the position estimation accuracy generally improves with an increasing number of particles, as expected. However, when the particle count is reduced to 1000, the conventional PF exhibits a notable decline in position estimation accuracy. In contrast, the proposed RBPF maintains position estimation accuracy despite the reduction in particle count. Furthermore, a comparison of the results between the proposed method and the conventional PF reveals a performance disparity of approximately 15\%, especially when the number of particles is set to 1000. This observation indicates that the proposed method can sustain high-precision position estimation even with a diminished number of particles. This improvement is attributed to the improved velocity estimation accuracy achieved through Rao-Blackwellization, which mitigates the likelihood of particle tracking failure. These results demonstrate that the proposed method is robust to changes in the number of particles and can maintain accuracy even in scenarios with limited computational resources. This renders the proposed method essential for real-world applications where real-time performance is critical.

\section{Discussion}
The evaluation results demonstrate that the proposed method enhances the velocity estimation accuracy by applying Rao-Blackwellization to the conventional PF and estimating velocity for each particle. This enhancement also increases the precision of state transitions within the PF, thereby resulting in improved position estimation accuracy with AFV, which does not necessitate ambiguity resolution. Furthermore, compared to the conventional PF and RTK-GNSS, the proposed method enables more stable position estimation in urban environments where GNSS signals are obstructed. Additionally, the proposed method maintains position estimation accuracy even with fewer particles, thereby allowing efficient position estimation in challenging urban environments where computational resources are constrained due to signal obstructions and require real-time processing.

\section{Conclusion}
In this study, we propose an accurate position estimation method using a RBPF that does not require integer ambiguity resolution of GNSS carrier phases. The proposed method incorporates Rao-Blackwellization into the PF framework established in our previous study. This modification facilitates the use of the KF for velocity estimation, which enhances precision of state transitions in the PF and improves position estimation accuracy in urban environments.

We employed GNSS data collected in urban environments to assess the accuracy of both velocity and position estimation, as well as the effects of varying the number of particles. The results confirmed that the proposed method achieves higher velocity estimation accuracy compared to conventional methods, enabling more accurate position estimation. Furthermore, the proposed method maintained its position estimation performance despite a reduced number of particles, thereby contributing to lower computational costs.

A future challenge arises as the proposed method currently relies on velocities calculated from GNSS Doppler observations for the linear components in the KF. However, when the number of available satellites is extremely limited, velocity cannot be computed, rendering observations for the linear components unavailable. To address this issue, we plan to implement a tightly coupled approach that directly utilizes GNSS Doppler measurements as observations for the linear components in the KF, thereby enhancing velocity estimation performance even with a limited number of GNSS satellites.

\bibliographystyle{IEEEtran.bst}
\balance
\bibliography{conference_101719}

\end{document}